\documentclass{article}

\usepackage{array}
\usepackage{makecell}
\usepackage{graphicx}
\usepackage{multirow}
\usepackage{arxiv}

\usepackage{float}
\usepackage{booktabs}
\usepackage{placeins}

\usepackage{geometry}
\usepackage[T1]{fontenc}
\usepackage{amsfonts}
\usepackage{fix-cm}
\usepackage{nicefrac}
\usepackage{microtype}

\usepackage{hyperref}
\usepackage{url}

\usepackage{xcolor}
\graphicspath{{./images/}}

\title{GSM-Plus-BN: A Perturbation-Based Benchmark for Bangla Mathematical Reasoning in Large Language Models}

\author{
  Bidyarthi Paul \\
  Computer Science and Engineering \\
  Southeast University \\
  Dhaka, Bangladesh \\
  \texttt{bidyarthipaul01@gmail.com}
  \And
  Nahida Jannat Mayouree \\
  Computer Science and Engineering \\
  Southeast University \\
  Dhaka, Bangladesh \\
  \texttt{nahidajannatmayouree@gmail.com}
  \And
  Md. Asif Karim \\
  Computer Science and Engineering \\
  Southeast University \\
  Dhaka, Bangladesh \\
  \texttt{mdasifkarim28th@gmail.com}
  \And
  Sagar Chandra Nath \\
  Computer Science and Engineering \\
  Southeast University \\
  Dhaka, Bangladesh \\
  \texttt{sagarcnath@gmail.com}
  \And
  Swastika Kundu \\
  Computer Science and Engineering \\
  Ahsanullah University of Science and Technology \\
  Dhaka, Bangladesh \\
  \texttt{swastikakundu123@gmail.com}
}

\begin{document}
\maketitle

\begin{abstract}
The evaluation of mathematical reasoning in large language models (LLMs) has predominantly focused on high-resource languages like English. This has created a significant barrier to the equitable development and deployment of AI in linguistically diverse regions such as Bangladesh, where over 230 million people speak Bengali. Despite this global significance, there has been minimal prior work on mathematical reasoning in Bengali and no existing research that systematically benchmarks a perturbated Bengali mathematical dataset, leaving a critical void in assessing model robustness and true comprehension beyond pattern recognition. This study addresses this gap by introducing GSM-Plus-BN, a novel perturbated Bengali mathematical dataset derived from the English GSM-Plus benchmark and verified by human translators. We evaluate six open-source LLMs Qwen3-32B, Llama-3.1-8B-Instant, Llama-3.3-70B-Versatile, Llama-4-Scout-17B-16E-Instruct, GPT-OSS-120B, and GPT-OSS-20B using a benchmark of 9,000 evaluation samples comprising 1,000 seed questions and 8,000 perturbed variants under both Standard Prompting and Chain-of-Thought (CoT) Prompting. Experimental results show that GPT-OSS-20B achieves the highest seed question accuracy of 96.08\% under Standard Prompting, while larger models such as Llama-3.3-70B and GPT-OSS-120B demonstrate superior robustness across perturbation types. Furthermore, CoT prompting substantially improves reasoning for most models compared to Standard Prompting, yet a notable performance gap persists across all models relative to their English benchmarks, underscoring the inherent difficulty of perturbed Bengali text. This research makes a foundational contribution by providing GSM-PLUS-BN as a new resource and baseline for future Bengali mathematical reasoning research.
\end{abstract}



\section{Introduction}

Mathematical reasoning has emerged as one of the most fundamental and challenging capabilities for evaluating the cognitive abilities of Large Language Models (LLMs)~\cite{cobbe2021training,hendrycks2021measuring,brown2020language,chowdhery2022palm,rae2021scaling}. Unlike simple fact recall or basic language understanding, mathematical word problem solving requires a complex integration of multiple cognitive skills: natural language comprehension, logical deduction, numerical computation, multi-step planning, and the ability to maintain coherent reasoning chains over extended sequences~\cite{wei2022chain,lampinen2022languagemodelslearnexplanations, li2024evaluating}. These requirements make mathematical reasoning an ideal benchmark for assessing the depth and robustness of LLM capabilities, revealing not just what models know, but how they think~\cite{srivastava2023beyond,anand2024mathifyevaluatinglargelanguage,suzgun2022challenging}.

In recent years, the research community has developed numerous benchmarks to systematically evaluate mathematical reasoning \cite{rong2025benchmarking, gao2025omni, sessler2024benchmarking, mishra2022lila}. The GSM8K dataset \cite{cobbe2021training} introduced a collection of grade-school math problems that has become a standard reference point. The MATH dataset \cite{hendrycks2021measuring} expanded this by including more challenging problems across diverse mathematical domains. Subsequently, datasets such as LILA \cite{mishra2022lila}, MathVista \cite{lu2024mathvista}, SVAMP \cite{patel2021nlp}, FrontierMath \cite{glazer2025frontiermathbenchmarkevaluatingadvanced}, and MMLU-Math \cite{hendrycks2020measuring}, MATH-Perturb \cite{huang2025mathperturbbenchmarkingllmsmath} and Omni-math \cite{gao2025omni} have further enriched the evaluation landscape, each targeting different aspects of mathematical reasoning such as robustness to numerical variations, program synthesis, and multi-domain coverage \cite{zou2025dynamath}.

A significant advancement came with the introduction of GSM-Plus \cite{li2024gsm}, a benchmark specifically designed to evaluate model robustness through linguistic perturbations. Unlike traditional benchmarks that measure only accuracy on static problem formulations, GSM-Plus systematically tests model resilience across multiple perturbation types. The authors identify five perspectives to guide the development of GSM-Plus: (1) Numerical Variation (numerical substitution, digit expansion, and integer-decimal-fraction conversion), (2) Arithmetic Variation (adding operation and reversing operation), (3) Problem Understanding (rephrasing text descriptions), (4) Distractor Insertion (inserting topic-related but useless sentences), and (5) Critical Thinking (focusing on question or doubt ability when the question lacks necessary statements).

The importance of robustness evaluation has been increasingly recognized as LLMs are deployed in real-world applications where input variations are inevitable \cite{yang2025evaluating, hao2025investigationrobustnessllmsmathematical}. Users naturally express problems in diverse ways, and models that are brittle to paraphrasing or minor structural changes are unlikely to provide reliable service in production environments \cite{kumar2025robustness, ackerman2024novel}. Furthermore, robustness to perturbations serves as a proxy for true understanding: models that merely memorize problem patterns rather than genuinely comprehend mathematical structures will fail when those patterns are disrupted \cite{venkatasubramanian2025large, zhou2025your}.

\begin{figure}[h]
    \centering
    \includegraphics[width=1\textwidth]{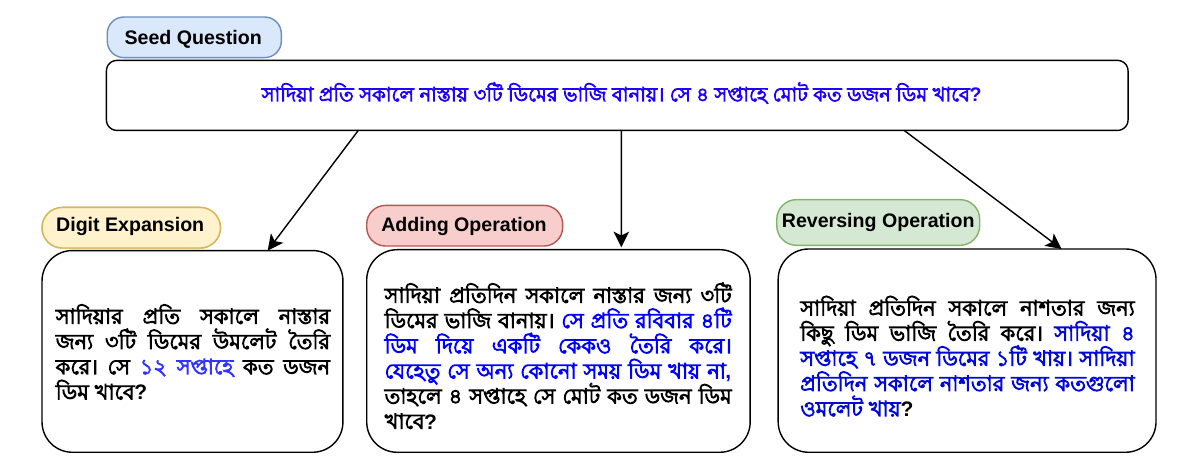}
    \caption{A seed example and its corresponding variants generated using three perturbation types.}
    \label{fig:intro-pdf}
\end{figure}

Despite the substantial progress in English-language mathematical reasoning evaluation, the research landscape for low-resource languages remains severely underexplored \cite{poria2025bhaasha, alam2024llms}. This disparity is not merely a matter of academic interest but has profound implications for the equitable development and deployment of AI technologies across global linguistic communities \cite{pakray2025natural2}. As LLMs increasingly power applications used worldwide, the lack of robust evaluation frameworks for languages other than English creates significant risks of biased performance and exclusion \cite{ki2025linguistic}. Bangla, the seventh most spoken language in the world with over 300 million native speakers primarily in Bangladesh and the Indian state of West Bengal, exemplifies this disparity\cite{wiki_bengali},\cite{paul2025geospatial}. Despite its vast speaker population, Bangla remains severely underrepresented in NLP research, particularly for complex reasoning tasks \cite{prama2025banglamath, paul2025leveraging}. This underrepresentation is striking when compared to high-resource languages like English, Chinese, or Spanish, which enjoy abundant datasets, models, and evaluation frameworks \cite{dipta2026ganitllm}.

The scarcity of Bangla resources extends across multiple dimensions of NLP. In terms of language models, while English boasts hundreds of pretrained models, Bangla has only a handful such as BanglaBERT \cite{bhattacharjee2022banglabert} and BanglaBERT-Base \cite{saha2023banglabert}. For benchmark datasets, initiatives like XTREME \cite{hu2020xtreme}, and XGLUE \cite{liang2020xglue} have included only a limited set of low-resource languages \cite{chen2024breaking}, with Bangla often absent or represented only in token amounts. For named entity recognition, the BanglaNER dataset \cite{islam2020banglaner} provides some coverage, but these are fundamentally different from the reasoning tasks we investigate. Recent work has introduced BnMMLU \cite{joy2026bnmmlu}, a benchmark to evaluate multitask language understanding in Bengali across 23 domains, revealing significant performance gaps that highlight the need for improved pre-training and fine-tuning strategies tailored to Bengali data.

The gap is particularly acute in mathematical reasoning. While English-language mathematical reasoning has been extensively studied through benchmarks like GSM8K \cite{cobbe2021training}, MATH \cite{hendrycks2021measuring}, GSM-Plus \cite{li2024gsm}, MathQA \cite{amini2019mathqa}, and others, the equivalent resources for Bangla are limited. Recent efforts have emerged, including BanglaMATH \cite{prama2025banglamath} and SOMADHAN \cite{paul2025leveraging}. However, these resources remain limited in scope and lack the comprehensive perturbation-based robustness evaluation that has become standard in English-language benchmarks.

Recognizing this critical gap, we introduce GSM-PLUS-BN, a Bangla-language counterpart to the English GSM-Plus dataset. This publicly available resource provides a parallel, perturbed mathematical reasoning benchmark for Bangla, enabling systematic cross-lingual evaluation of LLM robustness. The dataset comprises 10,544 perturbed question variations derived from 1,318 seed questions from GSM-Plus, incorporating eight distinct perturbation types: numerical substitution, digit expansion, integer-decimal-fraction conversion, adding operation, reversing operation, problem understanding, distraction insertion, and critical thinking. These perturbations are specifically designed to probe different aspects of model robustness, from handling numerical variations to resisting semantically coherent but irrelevant information. As shown in Figure \ref{fig:intro-pdf}, perturbations are applied to a seed example to produce altered versions that can be used to analyze model behavior under different input conditions. This paper conducts a comprehensive comparative evaluation of six diverse LLMs using the GSM-PLUS-BN dataset. The models in our study span a range of architectures and parameter scales, providing a representative cross-section of current LLM capabilities: Qwen3-32B \cite{yang2025qwen3,bai2023qwen}, Llama-3.1-8B-Instant, Llama-3.3-70B-Versatile, Llama-4-Scout-17B-16E-Instruct \cite{touvron2023llama,touvron2023llama,grattafiori2024llama,abdullah2025evolution}, GPT-OSS-120B, and GPT-OSS-20B \cite{agarwal2025gpt}. We evaluate all models using two standardized prompting strategies: standard prompting (direct problem presentation without examples) and chain-of-thought prompting (encouraging step-by-step reasoning) \cite{wei2022chain, kojima2023largelanguagemodelszeroshot, wang2023understandingchainofthoughtpromptingempirical}. This dual approach enables assessment of both base reasoning capabilities and the effectiveness of prompting techniques in Bangla.

Through this work, we make the following key contributions:

\begin{itemize}
    \item This paper introduces GSM-PLUS-BN, the first comprehensive perturbed mathematical reasoning benchmark for Bangla, comprising 10,544 question variations across eight perturbation types. 
    \item A detailed analysis of model performance across eight perturbation types is conducted, identifying which types pose the greatest challenges in the Bangla context. The analysis reveals how different perturbation mechanisms from numerical substitutions to distraction insertion affect model performance.
    \item Insights into how model scale and architectural choices affect performance in low-resource language settings are provided, offering practical guidance for model selection when deploying reasoning systems for Bangla.
    \item The effectiveness of standard prompting and chain-of-thought prompting strategies in Bangla is evaluated, revealing optimal prompting approaches for different model sizes and perturbation types.
\end{itemize}

\section{Related Work}

This section provides a comprehensive review of prior work, systematically analyzing recent foundational studies that collectively inform our understanding of mathematical reasoning in LLMs. With particular attention to prompting techniques, benchmark development, Bengali language resources, robustness evaluation, multilingual reasoning, multimodal reasoning, and advanced theorem proving. Table \ref{tab:related_work_expanded} presents a comprehensive summary of prior work on mathematical reasoning.

\subsection{Prompting Techniques for Mathematical Reasoning}

The foundational work of \cite{kojima2022zero} introduced Zero-shot Chain-of-Thought (Zero-shot-CoT) prompting, establishing that reasoning capabilities can be elicited by simply adding the phrase ``Let's think step by step'' to prompts. Their experiments demonstrated significant performance improvements across eight reasoning benchmarks without requiring examples or fine-tuning. The method proved particularly effective because it leverages the emergent reasoning capabilities of LLMs through a simple yet powerful prompting technique. However, the effectiveness depends substantially on model size, with smaller models showing limited improvements, suggesting that certain reasoning capabilities only emerge above specific parameter thresholds. Building upon this foundation, proposed MathPrompter \cite{imani2023mathprompter}, a framework that enhances mathematical reasoning by generating multiple solution paths using Zero-shot Chain-of-Thought prompting. Their approach achieved a notable accuracy improvement from 78.7\% to 92.5\% on the MultiArith dataset using a 175B parameter GPT-based model. 

However, \cite{imani2023mathprompter} acknowledged that both algebraic and Python-based solutions may still be incorrect if they share the same underlying reasoning error, highlighting the need for more robust and fundamental error-checking mechanisms. The extended prompting \cite{paul2025leveraging} research by investigating Tree-of-Thought (ToT) prompting for Bengali mathematical word problems from the SOMADHAN dataset, comparing standard prompting, Chain-of-Thought, and ToT strategies. Their work demonstrated that ToT achieved the highest accuracy of 88\% with GPT-OSS-120B and improved performance for medium-to-large models compared to CoT. However, smaller models showed unstable performance under ToT, and the study was limited to only 100 representative problems.

Conducted \cite{sessler2024benchmarking} a comprehensive evaluation of four large language models using five benchmark datasets and seven prompting strategies, finding that GPT-4o achieved the best overall results while larger models were less sensitive to prompting techniques than smaller models. Critically, they discovered that no single prompting strategy consistently outperformed others across all tasks, suggesting that optimal prompting approaches are task-dependent.

\subsection{Comprehensive Surveys on Mathematical Reasoning}

Provide \cite{wang2025survey} a comprehensive survey of large language models for mathematical reasoning, organizing prior work using a dual cognitive framework of comprehension and generation. Their review covers extensive datasets and methodologies including Chain-of-Thought prompting, supervised fine-tuning, and inference scaling, demonstrating that LLMs achieve strong performance across established benchmarks such as GSM8K and MATH. However, as a review paper, no single accuracy metric is reported, and the authors identified the lack of standardized evaluation as a major limitation, highlighting the urgent need for unified benchmarks to enable reliable cross-model assessment.

Present \cite{wang2026survey} a foundational and comprehensive analysis of LLMs in mathematical reasoning, systematically categorizing methodologies from basic prompting to active fine-tuning across diverse datasets. Their holistic review unifies a highly fragmented research landscape, demonstrating that models utilizing specialized mathematical tokenization and Program-of-Thought architectures achieve significantly superior accuracy and verifiable explainability. However, \cite{wang2026survey} identified inherent brittleness and limited generalization as persistent challenges, noting that performance remains highly inconsistent when models encounter adversarial phrasing or shifting grade levels, and that these models still lack human-centric pedagogical adaptability.

\subsection{Bengali Mathematical Reasoning Resources}

The development of resources for Bengali mathematical reasoning represents a recent but rapidly growing area of research, addressing a critical gap for the world's seventh most spoken language. Despite Bengali's significant global footprint, computational resources and benchmarks in the language have historically been scarce, creating substantial barriers for developing and evaluating AI systems tailored to Bengali-speaking populations.

Made \cite{paul2025leveraging} a seminal contribution by investigating the effectiveness of LLMs for solving Bengali math word problems using Chain-of-Thought reasoning on the SOMADHAN dataset. Their experiments demonstrated that LLaMA-3.3 70B achieved the best performance with 88\% accuracy using few-shot CoT prompting, demonstrating strong potential for multilingual mathematical reasoning. The dataset, containing 8,792 complex Bengali math word problems with manually written, step-by-step solutions, represents a significant advancement as no human-annotated Bengali dataset had previously addressed multi-step mathematical reasoning. However, \cite{paul2025leveraging} revealed limitations including a relatively small evaluation set and weaker performance of smaller models in understanding Bengali language structures, and noted that fine-tuning benefits were limited for smaller models, highlighting persistent challenges in low-resource language adaptation.

Introduced the BenNumEval \cite{ahmed2025bennumeval}, a Bengali numerical reasoning benchmark containing 3,255 samples across six task types including Causal Reasoning, Data Sufficiency, Cause and Effect Questions, Fill in the Blanks, Question-Answering Natural Language Inference, and Arithmetical Word Problems. Their evaluation using Bengali Native Prompting (BNaP), Cross-Lingual Prompting (XLP), and Cross-Lingual Chain-of-Thought (XCoT) strategies showed that Gemini 2.0 Flash achieved the best average accuracy of 79.63\% with XLP, while the human baseline reached 98.05\%. This substantial performance gap between state-of-the-art LLMs and human performance reveals significant challenges remaining in Bengali numerical reasoning. However, \cite{ahmed2025bennumeval} noted that the dataset size is relatively moderate and limited to Bengali, with mostly zero-shot evaluation, and some translated questions may introduce bias.

Contributed BanglaMATH \cite{prama2025banglamath}, a benchmark dataset of 1,700 Bangla math word problems for grades 6-8, carefully curated from Bangla school textbooks and exams with problems annotated by grade, reasoning steps, and digit count. Their evaluation revealed that DeepSeek V3 achieved the highest accuracy (86.9\% for Grade 6, 82.2\% for Grade 8), while Grok 3 performed the lowest. Critically, when augmenting problems with distracting information and translating them to English, both top-performing models demonstrated significant robustness failures and performance bias in Bangla, highlighting the vulnerability of LLMs to linguistic and contextual challenges. This paper \cite{prama2025banglamath} acknowledged limitations due to budget and API constraints, leaving many open-source and multilingual LLMs untested. Presented an end-to-end Bangla AI system for solving Math Olympiad problems \cite{ahmed2025bennumeval} by introducing the BDMO dataset and integrating fine-tuned LLMs with RAG and Tool-Integrated Reasoning (TIR). Their best-performing model achieved a score of 77/100 on the test set, demonstrating strong potential for Bangla numerical reasoning. However, \cite{ahmed2025bennumeval} noted that keyword-based RAG did not significantly improve performance and that models still perform weaker in Bangla compared to English, with a key limitation being the small size of curated Bangla benchmarks restricting deep reasoning and generalization.

Introduced the PatiGonit \cite{era2024empowering} dataset of 10,000 Bengali math word problems, fine-tuning transformer models including mT5, mBART50, BanglaT5, and a basic transformer to generate equations from text. Their approach achieved top accuracies of 97.30\% with mT5 and 97.20\% with mBART50 alongside high BLEU scores exceeding 94. However, \cite{era2024empowering} faced challenges in cultural translation nuances, dataset limitations to basic arithmetic without diverse multi-step problems, and resource constraints hindering full dataset expansion or exhaustive hyperparameter tuning. Introduced the BMWP \cite{mondal2025bmwp} dataset of 8,653 Bengali math word problems aimed at operation prediction and problem-solving, applying RNN-based models with GloVe embeddings and fine-tuned GPT-2, achieving the highest validation accuracy of 90.42\% with LSTM. However, \cite{mondal2025bmwp} noted that the dataset is limited to single-step arithmetic problems and lacks complex, multi-step, or real-world scenarios, and LLMs like GPT-2 or mT5 struggled in zero-shot settings.

Presented MathMist \cite{sobhani2026mathmist}, a parallel multilingual benchmark with 2,890 gold-standard Bangla-English math problems from secondary textbooks, expanded synthetically to 13 languages via LLMs, evaluating GPT-OSS under zero-shot, CoT, and code-switched paradigms, yielding strong Bangla performance of 88.97\% overall accuracy and slightly higher English results at 90.20\%. Despite these encouraging results, \cite{sobhani2026mathmist} identified limitations including restricted coverage to 13 languages excluding many low-resource ones, secondary-level difficulty without advanced topics, narrow error analysis, and metrics overlooking subtle reasoning flaws. Introduced the Bangla-Bayanno \cite{hasan2025bangla}, a Bengali Visual Question Answering dataset containing 52,650 question-answer pairs translated from English VQA v2 using ChatGPT-4 and a multi-stage validation process. Human evaluation showed that the proposed translation approach outperformed manual and conventional machine translation in preserving semantic quality. However, \cite{hasan2025bangla} noted that some translated terms remain language-dependent and visual diversity is limited.

Introduced MathQuest \cite{anand2024mathify}, a benchmark for evaluating mathematical reasoning using NCERT Class 11 and 12 mathematics problems, fine-tuning LLaMA-2, WizardMath, and MAmmoTH using QLoRA, with MAmmoTH-13B achieving the best performance. However, \cite{anand2024mathify} acknowledged that the benchmark is limited to school-level mathematics and a few open-source models. Introduced Distract Math-BN \cite{sobhani2026mathmist}, a Bangla benchmark augmenting MGSM and MSVAMP with semantically coherent but computationally irrelevant distractors, evaluating seven models from 3B to 12B parameters. Their study revealed substantial performance degradation, with standard models dropping up to 41 points and reasoning-specialized models declining 14-20 points despite consuming 5x more tokens. These findings underscore the vulnerability of even advanced models to adversarial modifications, particularly in low-resource language settings.

\subsection{Robustness and Perturbation in Mathematical Reasoning}

Evaluating the robustness of mathematical reasoning under adversarial and perturbed conditions has emerged as a critical research direction.Introduced the GSM-PLUS \cite{li2024gsm}, a robust benchmark designed to evaluate the mathematical reasoning abilities of LLMs under adversarial conditions. Their results demonstrate that although models like GPT-3.5-Turbo achieve around 61\% accuracy, performance drops significantly compared to original GSM8K, indicating limited robustness to small perturbations. This suggests that many LLMs rely on surface-level patterns rather than deep reasoning. \cite{li2024gsm} acknowledged that GSM-PLUS is derived only from modified GSM8K problems, lacking entirely new problem types and higher-level mathematical complexity. Proposed Math-RoB \cite{yu2025benchmarking}, a benchmark specifically designed to evaluate the robustness of mathematical reasoning by modifying problem formulations while preserving the reasoning process. Experimental results showed that all evaluated models experienced noticeable performance degradation, revealing fundamental weaknesses in reasoning robustness. \cite{yu2025benchmarking} recommended extending the benchmark to broader reasoning domains and developing more robust training strategies.

Introduced MATH-Perturb \cite{huang2025math}, a benchmark that applies numerical, semantic, and structural perturbations to assess reasoning stability. Experimental results showed that all models experienced performance degradation, with OpenAI o1 demonstrating the strongest robustness, and the largest accuracy drops occurring under semantic rewriting and problem restructuring. \cite{huang2025math} noted that the benchmark is limited to text-based mathematics and manually designed perturbations, recommending extension with multilingual datasets. Proposed FormalMATH \cite{yu2025formalmath}, a benchmark for evaluating formal mathematical reasoning using the Lean 4 theorem prover, containing 5,560 formally verified mathematical problems requiring machine-verifiable proofs. Results showed that OpenAI o1 achieved the highest proof success rate, although many generated proofs failed due to logical or syntactic errors. Investigated \cite{xie2025memorization} whether LLMs rely on reasoning or memorization by introducing the Knights \& Knaves, DynamicZebra, and LiMem benchmarks. Results showed that fine-tuned models memorized training data but still demonstrated improved reasoning on unseen tasks, concluding that memorization and reasoning complement each other.

\subsection{Multilingual Mathematical Reasoning}

Multilingual mathematical reasoning represents a frontier challenge in AI, as models must navigate linguistic diversity while maintaining mathematical competence across languages. Introduced PolyMath \cite{wang2026polymath}, a multilingual mathematical reasoning benchmark covering 18 languages and multiple difficulty levels, finding that Qwen-3-235B-A22B-Thinking achieved the highest performance while significant performance gaps persisted between high-resource and low-resource languages. The proposed MMATH \cite{luo2025mmath}, covering 10 languages with competition-level problems, identifying off-target language generation issues, and showing that reasoning in English before generating answers in the target language improved performance.

MathEval \cite{liu2025matheval}, which integrates 22 mathematical reasoning datasets in English and Chinese, shows that reasoning-specialized models outperformed general-purpose models. FrontierMath \cite{glazer2024frontiermath} for advanced mathematical reasoning using research-level problems, showing that OpenAI o1-preview achieved the highest accuracy at 25.2\% while most other models solved fewer than 2\% of problems. Introduced OMNI-MATH \cite{gao2025omni} containing 4,428 Olympiad-level mathematics problems, showing that OpenAI o1-preview achieved approximately 58\% accuracy but all models remained well below expert human performance. Introduced LILA\cite{mishra2022lila}, a unified benchmark integrating 23 mathematical reasoning tasks, and proposed the BHASKARA multi-task learning model achieving a 21.83\% improvement in F1 score compared to task-specific models.

\begin{table*}[htbp]
\centering
\caption{Summary of Related Work on Bengali Mathematical Reasoning and Perturbation}
\label{tab:related_work_expanded}

\fontsize{11pt}{13pt}\selectfont
\renewcommand{\arraystretch}{1.25}
\setlength{\tabcolsep}{3.8pt}

\resizebox{\textwidth}{!}{%
\begin{tabular}{|>{\arraybackslash}p{4.6cm}|
                >{\arraybackslash}p{3.5cm}|
                >{\centering\arraybackslash}p{2.5cm}|
                >{\centering\arraybackslash}p{2.2cm}|
                >{\centering\arraybackslash}p{2.1cm}|
                >{\centering\arraybackslash}p{1.8cm}|}
\hline

\centering \textbf{Paper Title} &
\centering \textbf{Approach} &
\textbf{Model} &
\textbf{Dataset} &
\textbf{Accuracy} &
\textbf{Language} \\
\hline

GSM-PLUS: Evaluating Mathematical Reasoning Robustness &
Adversarial problem modifications; perturbed GSM8K &
GPT-3.5-Turbo &
GSM-PLUS &
$\sim$61\% &
English \\
\hline

Math-RoB: Robustness Benchmark with Mathematically-Equivalent Transformations &
Linguistic and parametric variations on advanced problems &
OpenAI O3 &
PutnamGAP &
Dropped 4.7--12.9 pts &
English \\
\hline

MATH-Perturb: Numerical, Semantic, Structural Perturbations &
Three perturbation types (numerical, semantic, structural) &
OpenAI o1 (most robust) &
MATH-Perturb &
Performance degraded &
English \\
\hline

SOMADHAN: Bengali Math Word Problems with CoT &
Few-shot Chain-of-Thought prompting &
LLaMA-3.3 70B &
SOMADHAN &
88\% &
Bengali \\
\hline

PatiGonit: Bengali Math Word Problems &
Fine-tuning transformers (mT5, mBART50, BanglaT5) &
mT5 &
PatiGonit &
97.30\% &
Bengali \\
\hline

Tree-of-Thought Prompting for Bengali Math &
ToT vs.\ CoT vs.\ standard prompting &
GPT-OSS-120B &
SOMADHAN &
88\% &
Bengali \\
\hline

PolyMath: Multilingual Benchmark Covering 18 Languages &
Multiple difficulty levels; cross-lingual performance analysis &
Qwen-3-235B &
PolyMath &
Gaps across languages &
18 Languages \\
\hline

MathMist: Parallel Multilingual Benchmark with Perturbation Reasoning &
Zero-shot, CoT, code-switched, and perturbed reasoning &
GPT-OSS-20B &
MathMist &
88.97\% (Bangla) &
13 Languages \\
\hline

\end{tabular}%
}

\end{table*}

\subsection{Multimodal Mathematical Reasoning}

The integration of visual and textual information for mathematical reasoning has emerged as another important research direction. \cite{lu2024mathvista} developed the MathVista benchmark, aggregating 6,141 multimodal math problems from 28 datasets, evaluating 12 foundation models, with GPT-4V leading at 49.9\% accuracy. Introduced the MATH-Vision \cite{wang2024measuring} dataset with 3,040 visually grounded problems from real math competitions, showing that current models still perform significantly below human level especially in complex domains like geometry and topology. The proposed Of MultiMath \cite{peng2024multimath}, a multimodal framework achieving 58.5\% accuracy on MathVista and 48.0\% on MathVerse, though reporting limited generalization to out-of-domain datasets.

\subsection{Advanced Mathematical Reasoning and Fine-Tuning}

Introduced the DeepSeekMath \cite{shao2024deepseekmath}, an open-source mathematical reasoning framework trained on a large-scale corpus, achieving nearly 60\% accuracy and supporting both English and Chinese benchmarks. However, \cite{shao2024deepseekmath} noted that DeepSeekMath still lags behind GPT-4 in few-shot generalization. Investigated advancing math \cite{chen2025advancing} reasoning in 8B Llama-based models through continued pretraining with problem-solving data versus general corpora, yielding MathGPT-8B with superior GSM8K (64.44\%) and MATH (35.88\%) scores. Identified \cite{chen2025advancing} that supervised fine-tuning was inferior to continued pretraining due to poor out-of-distribution handling and hard data adaptation. Introduced the CARP \cite{zhang2023evaluating} benchmark to systematically evaluate computation-intensive mathematical reasoning in tool-augmented language models, achieving 73.46\% accuracy on the CARP dataset using their DELI approach.

However, \cite{zhang2023evaluating} noted limitations in the scale of the dataset and restricted access to certain models. Proposed MARIO \cite{liao2024mario}, a reproducible pipeline that integrates REACT-style prompting with code interpreter outputs, achieving 69.69\% accuracy on GSM8K. However, they identified data quality and format inconsistencies along with the need for human verification as key limitations. Introduced MALT \cite{motwani2024malt}, a multi-agent post-training framework that teaches specialized LLMs to collaborate via tree-search and value-iteration, achieving relative improvements of 15.66\% on MATH, 9.40\% on CSQA, and 7.42\% on GSM8K. However, the approach is constrained by severe computational intensity and is currently limited to deterministic domains.

\subsection{Research Gap}

Based on our systematic analysis of reviewed papers, we identify two critical research gaps that motivate our study:

First, there has been not too much work done on Bengali language or Bengali Math before. While papers by \cite{paul2025leveraging}, \cite{ahmed2025bennumeval}, \cite{prama2025banglamath}, \cite{era2024empowering}, \cite{mondal2025bmwp}, \cite{sobhani2026mathmist}, \cite{hasan2025bangla}, \cite{ahmed2025bennumeval}, and \cite{nazi2026dag} have made foundational contributions to Bengali mathematical reasoning, the benchmarks remain limited in scale, scope, and problem diversity. Most existing Bengali datasets are restricted to basic arithmetic \cite{era2024empowering, mondal2025bmwp}, school-level mathematics \cite{prama2025banglamath, anand2024mathify}, or small evaluation sets \cite{paul2025leveraging, mahmood2025structured}. The Bengali mathematical reasoning landscape remains significantly under-resourced compared to English and other high-resource languages.

Second, and critically, there has been no work done on perturbed Bengali Math before. While robustness and perturbation research has progressed significantly in English through benchmarks like GSM-PLUS \cite{li2024gsm}, Math-RoB \cite{yu2025benchmarking}, and MATH-Perturb \cite{huang2025math}, and while Bengali robustness has been preliminarily explored through distractors \cite{nazi2026dag}, no study has systematically translated, evaluated, and benchmarked adversarial or perturbed mathematical reasoning in Bengali. The intersection of Bengali language and perturbed mathematical reasoning-specifically, how LLMs respond to numerical, semantic, and structural perturbations in Bengali-remains entirely unexplored.

\section{Dataset}
\begin{figure}[h]
    \centering
    \includegraphics[width=1\textwidth]{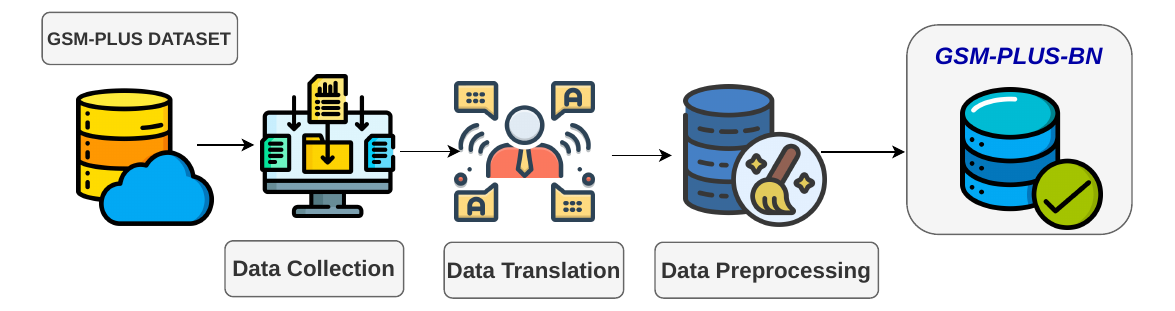}
    \caption{ Pipeline for the development of the GSM-PLUS-BN dataset}
    \label{fig:Dataset pipeline-pdf}
\end{figure}

To systematically evaluate the mathematical reasoning robustness of Large Language Models (LLMs) in Bengali, we introduce \textbf{GSM-PLUS-BN} adaptation of the GSM-PLUS dataset. Built upon the widely-adopted GSM8K benchmark \cite{cobbe2021training}, GSM-PLUS-BN comprises eight perturbation categories spanning five distinct dimensions. These perturbations are designed to probe specific vulnerabilities in LLM reasoning capabilities, ranging from numerical manipulation to logical fallacies. Figure~\ref{fig:perturbation} provides a comprehensive overview of all perturbation types with representative examples. Figure~\ref{fig:Dataset pipeline-pdf} provides a comprehensive overview of the development of our proposed dataset.

\subsection{Perturbation Taxonomy}
The GSM-Plus-BN dataset employs a systematic perturbation framework designed to evaluate the robustness and reasoning consistency of large language models (LLMs) in Bengali mathematical reasoning tasks. Each seed question serves as the base problem, from which eight distinct perturbation types are generated. These perturbations modify the surface form, numerical values, structure, or contextual elements of the problem while preserving the underlying mathematical logic and the correct answer. This controlled variation enables researchers to determine whether models genuinely understand mathematical concepts or rely on superficial pattern recognition.

\subsubsection{Numerical Variation}

This perturbation family examines whether LLMs rely on superficial numerical patterns rather than genuine mathematical reasoning. We implement three transformation strategies:

\textbf{Numerical Substitution} replaces numerical values with different numbers while preserving the same digit length. For example, a value of \textit{2} may be replaced with \textit{3} or \textit{4}. This tests whether models memorize specific numerical relationships or truly understand the underlying arithmetic operations.

\textbf{Digit Expansion} increases the number of digits while maintaining the same mathematical relationship. For example, \textit{2} becomes \textit{200} or \textit{2,000}. This perturbation evaluates whether models remain robust when numerical scales change.

\textbf{Integer-Decimal-Fraction Conversion} changes the representation of numerical values across equivalent or related formats. For example, an integer such as \textit{2} may be represented as \textit{2.0}, while other values may appear in decimal or fractional form (e.g., \textit{0.5} or \textit{1/2}). This evaluates whether models can correctly interpret different numerical representations.

\textit{Example Sequence:}
\begin{itemize}
    \item \textit{Seed:} A shirt requires 2 bolts of blue fiber and half as many bolts of white fiber. How many bolts of fiber are needed in total?
    \item \textit{Numerical Substitution:} A shirt requires 3 bolts of blue fiber and half as many bolts of white fiber. How many bolts are needed in total?
    \item \textit{Digit Expansion:} A shirt requires 200 bolts of blue fiber and half as many bolts of white fiber. How many bolts are needed in total?
    \item \textit{Conversion:} A shirt requires 2 bolts of blue fiber and 0.5 times as many bolts of white fiber. How many bolts are needed in total?
\end{itemize}

\subsubsection{Arithmetic Variation}

These perturbations evaluate whether models can flexibly apply arithmetic reasoning under modified problem structures.

\textbf{Adding Operation} introduces an additional arithmetic step that must be incorporated into the solution. This tests whether models can correctly perform multi-step reasoning without losing track of the original objective.

\textbf{Reversing Operation} converts the problem into its inverse form. Instead of asking for the final result given the inputs, the problem provides the result and asks the model to infer one of the original inputs. This evaluates bidirectional mathematical reasoning.

\textit{Example Sequence:}
\begin{itemize}
    \item \textit{Seed:} A shirt requires 2 bolts of blue fiber and half as many bolts of white fiber. How many bolts of fiber are needed in total?
    \item \textit{Adding Operation:} A shirt requires 2 bolts of blue fiber and half as many bolts of white fiber. Each bolt costs \$12. What is the total cost?
    \item \textit{Reversing Operation:} A shirt requires some blue fiber and half as much white fiber. The total number of bolts used is 3. How many bolts of blue fiber are required?
\end{itemize}

\subsubsection{Problem Understanding}

This category investigates whether models are sensitive to changes in wording while preserving the underlying mathematical meaning. We reformulate problems using different vocabulary and sentence structures without altering the required reasoning. This evaluates whether models genuinely understand problem semantics rather than relying on memorized textual patterns.

\textit{Example:}
\begin{itemize}
    \item \textit{Seed:} A shirt requires 2 bolts of blue fiber and half as many bolts of white fiber. How many bolts of fiber are needed in total?
    \item \textit{Problem Understanding:} If a shirt requires 2 units of blue material and half as many units of white material, how many units of material are required altogether?
\end{itemize}

\subsubsection{Distractor Insertion}

We insert topic-relevant but mathematically irrelevant statements that often include additional numerical values. This perturbation evaluates whether models can distinguish essential information from distracting details, a key requirement for solving real-world mathematical problems containing extraneous information.

\textit{Example:}
\begin{itemize}
    \item \textit{Seed:} A shirt requires 2 bolts of blue fiber and half as many bolts of white fiber. How many bolts of fiber are needed in total?
    \item \textit{Distractor Insertion:} A shirt requires 2 bolts of blue fiber and half as many bolts of white fiber. Each bolt normally costs \$5, but today there is a special discount that reduces the price of each bolt by \$5. How many bolts of fiber are needed to make the shirt?
\end{itemize}

\subsubsection{Critical Thinking}

Inspired by Polya's problem-solving principles, this perturbation removes essential information from the problem statement. Instead of generating an unsupported answer, models are expected to recognize that the available information is insufficient and explicitly identify the missing information. This evaluates higher-order reasoning and resistance to producing unjustified responses \cite{wei2022chain}.

\textit{Example:}
\begin{itemize}
    \item \textit{Seed:} A shirt requires 2 bolts of blue fiber and half as many bolts of white fiber. How many bolts of fiber are needed in total?
    \item \textit{Critical Thinking:} A shirt requires 2 bolts of blue fiber and some amount of white fiber. How many bolts are needed in total?
\end{itemize}

\subsection{Data Translation}

In this section, we describe the dataset translation process for GSM-PLUS-BN. We employed a hybrid approach combining AI-based tools with expert human refinement. An automated translation system first generated initial Bengali versions of all mathematical problems from the original GSM-PLUS dataset, which is in English. Subsequently, expert translators with strong backgrounds in both linguistics and mathematics manually reviewed and corrected each output. They followed strict guidelines to preserve mathematical expressions, contextual nuances, and step-by-step reasoning chains. This two-stage methodology effectively mitigated automated translation errors while ensuring logical coherence. The final dataset achieves high linguistic fluency and retains complete reasoning fidelity for reliable model evaluation.

\subsection{Translator Demographics and Expertise}

We engaged six skilled English-to-Bengali translators with expertise in both linguistics and mathematical reasoning. The entire corpus was evenly distributed among the translators to ensure efficiency and maintain uniform quality across the translated content. Table~\ref{tab:translators} provides detailed information about their backgrounds and experience. The team comprised one graduate, one postgraduate, and four undergraduate students, all specializing in Natural Language Processing (NLP), with ages spanning 22 to 30 years. Their research experience ranged from 1 to 4 years, with senior members overseeing the work of junior translators to ensure translation quality and consistency. This structured expertise distribution enabled effective quality control and maintained linguistic fidelity throughout the dataset refinement process.

Together, this combination of expert translation and multi-stage human refinement ensures that GSM-PLUS-BN maintains both linguistic fluency and logical coherence, making it a reliable benchmark for evaluating Bengali mathematical reasoning in large language models.

\begin{table}[htbp]
\centering
\caption{Detailed Information of Translators qualifications and professional experience}
\label{tab:translators}
\begin{tabular}{l c c c c c c}
\hline
\textbf{Details} & \textbf{T1} & \textbf{T2} & \textbf{T3} & \textbf{T4} & \textbf{T5} & \textbf{T6} \\
\hline
Role & Graduate & Graduate & Under-graduate & Under-graduate & Under-graduate & Under-graduate \\
Age & 30 & 23 & 24 & 24 & 22 & 25 \\
Research Field & NLP & NLP & NLP & NLP & NLP & NLP \\
Experience & 4 years & 2 years & 1 year & 1 years & 1 year & 1 years \\
\hline
\end{tabular}
\end{table}

\subsubsection{Translation Guidelines}

\begin{figure}[htbp]
    \centering
    \includegraphics[width=1.05\textwidth]{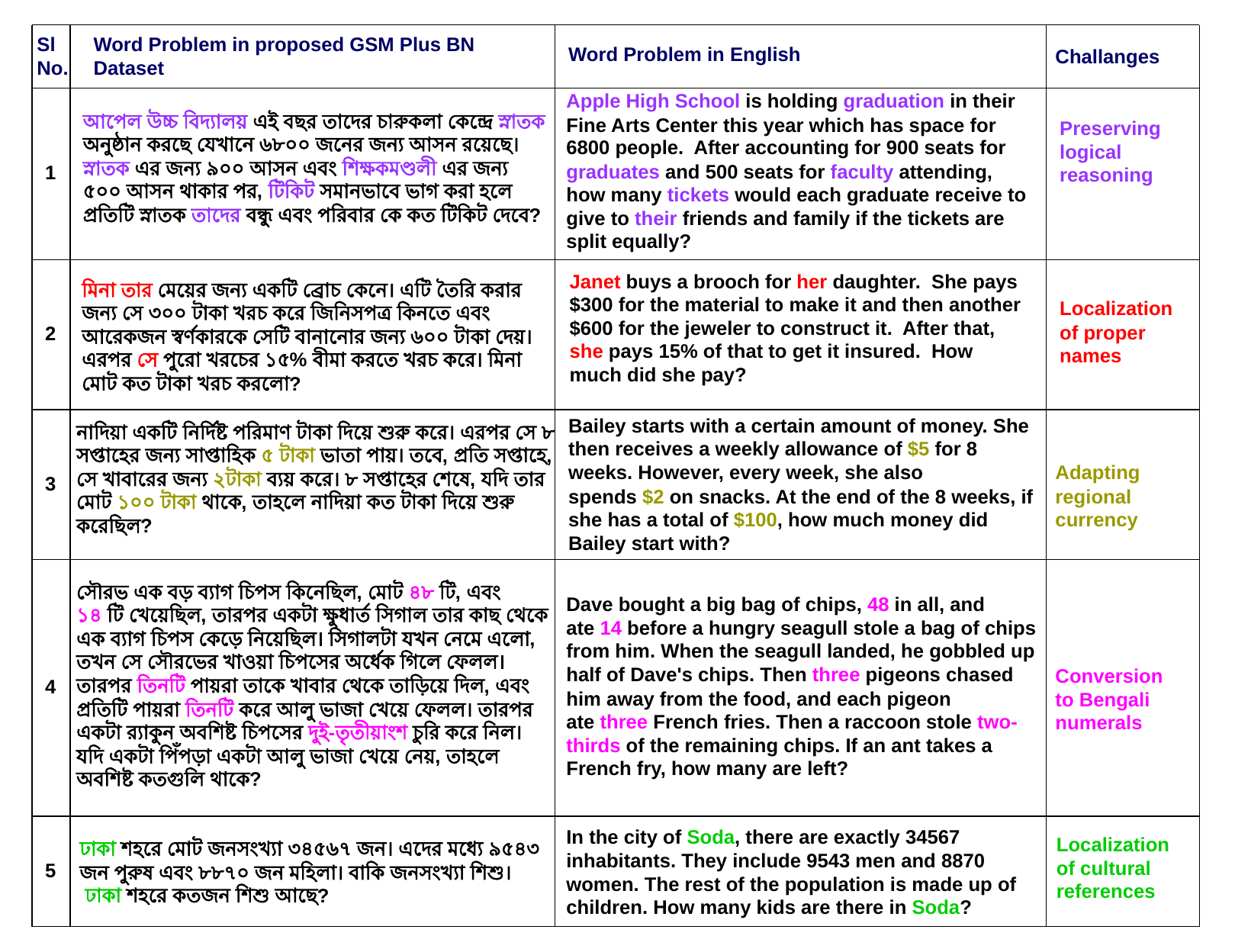}
    \caption{Challenges we faced during the translation of GSM-PLUS dataset}
    \label{fig:challenges}
\end{figure}

To guarantee translation quality, cultural relevance, and linguistic consistency, the following principles were established and communicated to the translators prior to the translation process. Figure~\ref{fig:challenges} provides an overview of the challenges encountered during translation.

\begin{enumerate}
    \item Translations preserve the original reasoning process and logical structure of each problem. Mathematical relationships, problem constraints, and contextual information remain unchanged to ensure semantic equivalence with the source text.
    
    \item Personal names are localized using culturally appropriate Bengali names to improve familiarity and naturalness for native readers. For example, the English name ``Janet'' is translated as ``Mina''.
    
    \item Currency references are adapted to the local context by replacing the Dollar symbol (\$) with the Bangladeshi Taka (BDT). For example, ``\$100'' is translated as ``100 BDT''.
    
    \item Arabic numerals are converted to Bengali numerals throughout the translated text to maintain orthographic consistency with the Bengali writing system.
    
    \item Location names and culturally specific entities are replaced with locally familiar equivalents whenever appropriate. These substitutions preserve the original problem structure and intended meaning while improving cultural relevance. For example, a foreign city name may be replaced with ``Dhaka,'' the capital of Bangladesh.
\end{enumerate}

\subsection{Dataset Characteristics}

The GSM-Plus-BN dataset comprises a total of 10,544 problem instances in Bengali. These instances were generated from 1,318 seed questions from the GSM8K test set. Following the original GSM-Plus methodology, each seed question was expanded into eight distinct adversarial variants. These variants preserve the correct numerical answer of the original seed question while introducing linguistic or structural perturbations designed to challenge model robustness without altering mathematical correctness.

\subsection{Data Structure}

 GSM-Plus-BN is organized into seven following structured columns:

\begin{itemize}
    \item \textbf{question:} The Bengali text of the perturbed math word problem. This is the input that models are expected to reason over, incorporating one of the eight adversarial perturbations while maintaining mathematical correctness.

    \item \textbf{solution:} The step-by-step reasoning chain and final answer derivation for the perturbed question, provided as a complete solution trace for training or evaluation purposes.

    \item \textbf{answer:} The final numerical answer to the perturbed problem, serving as the primary metric for evaluating model correctness.

    \item \textbf{perturbation\_type:} The specific category of adversarial modification applied to the seed question, enabling fine-grained analysis of model performance across different perturbation types.

    \item \textbf{seed\_question:} Each seed question serves as the base question from which all perturbations are generated, ensuring traceability to the original source.

    \item \textbf{seed\_solution:} The step-by-step reasoning solution for the original seed question, enabling comparative analysis between perturbed and unperturbed versions.

    \item \textbf{seed\_answer:} The final answer of the original seed question, which remains identical across all eight adversarial variants to ensure perturbations do not alter the underlying mathematics.
\end{itemize}







\subsection{Summary of Perturbation Examples}

\begin{figure}[htbp]
    \centering
    \includegraphics[width=1\textwidth]{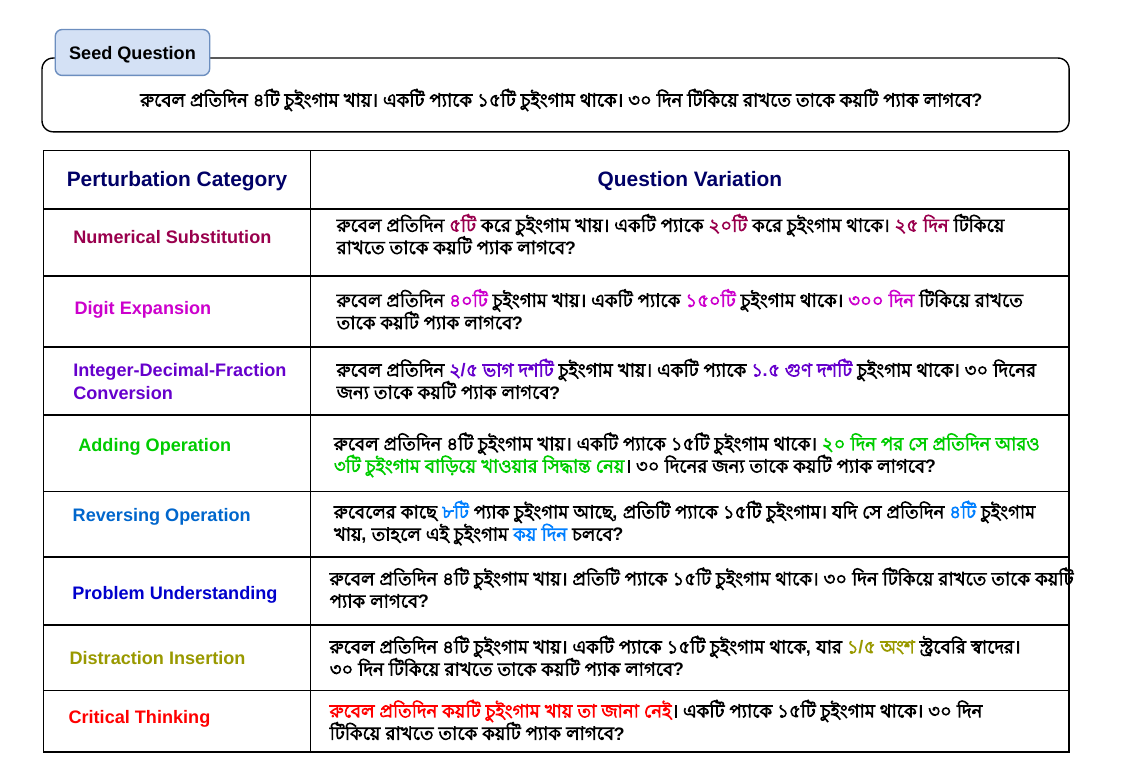}
    \caption{Summary of all perturbation types with Bengali examples}
    \label{fig:perturbation}
\end{figure}

Both GSM-Plus and GSM-Plus-BN employ the same adversarial perturbation framework to evaluate the robustness of large language models (LLMs) in mathematical reasoning. Since GSM-Plus-BN is a direct Bengali translation of GSM-Plus, the perturbation structure and reasoning logic remain identical across both datasets, enabling consistent cross-lingual evaluation. Figure~\ref{fig:perturbation} provides a overview of the Summary of all perturbation types with Bengali examples.

The perturbations modify the surface form of a problem while preserving its underlying mathematical semantics and correct answer. These variations include numerical changes, arithmetic operation modifications, paraphrasing for problem understanding, distractor insertion, and critical reasoning prompts. Such controlled variations help assess whether models rely on genuine reasoning rather than memorized patterns.

By generating multiple semantically equivalent variants of each seed problem, the perturbation framework enables systematic analysis of reasoning consistency across different linguistic and structural forms. Maintaining identical perturbation strategies across the English and Bengali datasets allows our study to directly compare model robustness and reasoning performance in both languages.

\subsection{Dataset Availability}

The GSM-Plus-BN dataset is publicly available for the research community. It is hosted on the Mendeley Data repository, and this release represents the third Version (V3) of the dataset. The GSM-Plus-BN dataset is openly available in Mendeley Data at  
\url{https://data.mendeley.com/datasets/74dscnmrhv/3}

\section{Methodology}

\begin{figure}[h]
    \centering
    \includegraphics[width=1\textwidth]{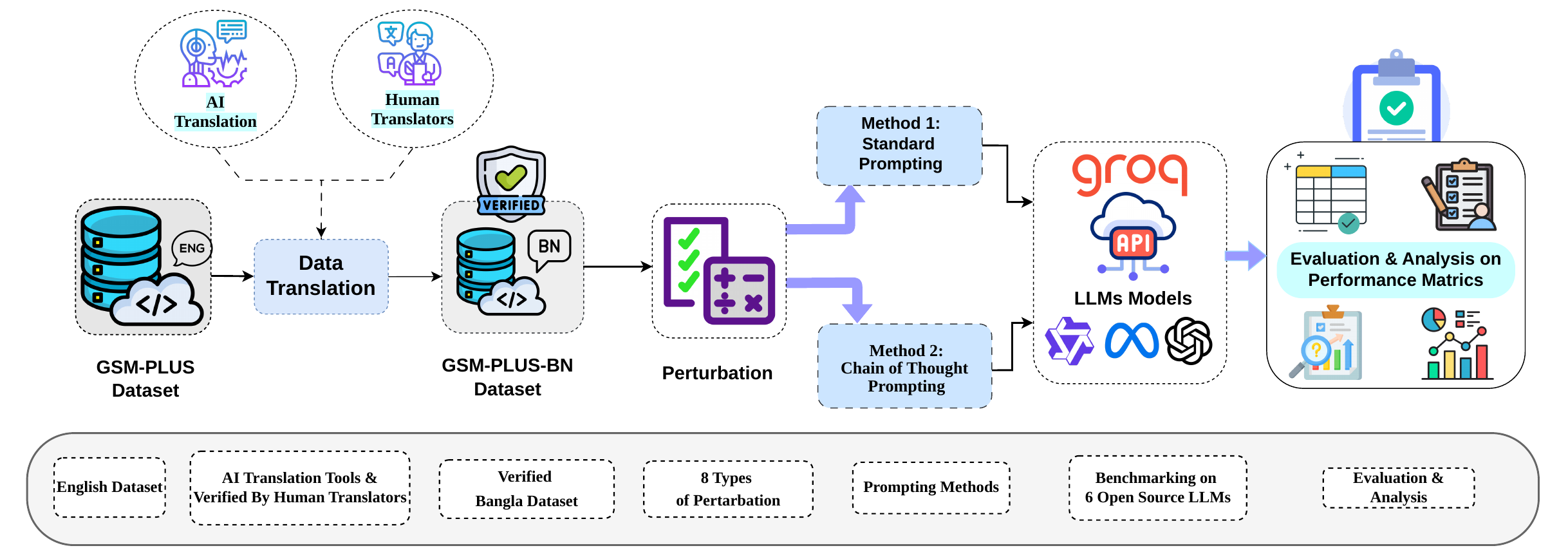}
    \caption{Schematic diagram of our proposed methodology.}
    \label{fig:Methodology-pdf}
\end{figure}

In this study, we present a comprehensive evaluation framework for benchmarking the mathematical reasoning capabilities of six open-source Large Language Models (LLMs) on perturbated Bengali mathematical problems. The framework evaluates model performance under two prompting strategies: Standard prompting and Chain-of-Thought (CoT) prompting. Each model is tested on the GSM-Plus-BN dataset using identical experimental settings to ensure a fair and reproducible ablation study. Figure~\ref{fig:Methodology-pdf} displays the schematic diagram of our proposed methodology.

\subsection{Selected Large Language Models}

In this study, we evaluate six open-source Large Language Models (LLMs) available through the Groq API platform \cite{groq2025api}. The selected models include both dense transformer and Mixture-of-Experts (MoE) architectures developed by Alibaba, Meta, and OpenAI. All models are instruction tuned and support multilingual reasoning, making them suitable for evaluating mathematical reasoning in both English and Bangla. The models were selected based on three primary criteria:

\begin{enumerate}
    \item \textbf{Open Source and Free Accessibility:} All selected models are open-source and freely available to the research community, ensuring reproducibility and accessibility for future research. The models can be accessed without cost through the Groq API, making them ideal for academic research with limited budgets.
    
    \item \textbf{Parameter Range (8B to 120B):} We selected models spanning a wide parameter range from 8 billion to 120 billion parameters. This diversity allows us to investigate how model size influences performance on perturbated Bengali mathematical reasoning tasks. Smaller models (8B-20B) help us understand the capabilities and limitations of computationally efficient models, while larger models (70B-120B) serve as performance benchmarks.
    
    \item \textbf{Suitability for Bengali Mathematical Reasoning:} All selected models support multilingual reasoning capabilities, making them suitable for evaluating mathematical reasoning in Bengali. The models represent different architectural families, including dense transformer and Mixture-of-Experts (MoE) architectures, enabling a comprehensive analysis of how architectural design influences reasoning performance.
\end{enumerate}

All models were evaluated using API-based inference through the Groq platform without additional fine-tuning or parameter modification. The following subsections provide detailed descriptions of each selected model.
To ensure a fair comparison, all models were evaluated using API-based inference without additional fine-tuning or parameter modification. Table~\ref{tab:models} summarizes the selected models and the reasons for their inclusion in this study.

\subsubsection{Qwen3-32B}

Qwen3-32B is an open-source large language model developed by Alibaba's Qwen team and released under the Apache 2.0 license \cite{qwen2025technical}. The model is designed for multilingual instruction following, logical reasoning, and complex problem solving, demonstrating strong performance across a wide range of language understanding and mathematical reasoning benchmarks, including GSM8K, MATH, AIME, and MGSM. The model supports more than one hundred languages, making it well suited for multilingual evaluation tasks. In this study, Qwen3-32B was accessed through the Groq API, which provides efficient inference for large-scale experiments. The model was selected because it offers a strong balance between reasoning capability, multilingual understanding, and computational efficiency. With 32 billion parameters, it represents a mid-range model in our evaluation, allowing us to examine how moderate parameter scales perform on perturbated Bengali mathematical reasoning tasks. Its proven ability to handle cross-lingual reasoning and instruction-following tasks makes it particularly suitable for evaluating mathematical reasoning in Bengali. Furthermore, as an open-source model freely available through the Groq API, it enables reproducible research without financial barriers, which is essential for academic studies with limited resources.

\subsubsection{Llama-3.1-8B-Instant}

Llama-3.1-8B-Instant is an open-source instruction-tuned language model developed by Meta AI and released under an open-source license \cite{meta2025llama31}. The model has been optimized for efficient inference while maintaining competitive reasoning and instruction-following capabilities. The model supports multilingual text generation and performs well on a variety of natural language processing tasks, including question answering, summarization, and mathematical reasoning. In this work, Llama-3.1-8B-Instant was accessed through the Groq API to ensure consistent and low-latency inference. The model was selected as the smallest model in our evaluation with 8 billion parameters to serve as a lightweight baseline. Including this model allows us to investigate the minimum parameter threshold required for reasonable performance on perturbated Bengali mathematical reasoning tasks. Its efficient computational requirements make it a practical choice for real-world applications where hardware resources and inference speed are important considerations. As an open-source model freely available through the Groq API, it provides an accessible entry point for researchers and developers working with limited computational resources. Comparing its performance with larger models provides valuable insight into the relationship between model size and reasoning ability in the context of Bengali perturbated mathematical problems.

\subsubsection{Llama-3.3-70B-Versatile}

Llama-3.3-70B-Versatile is a high-capacity open-source instruction-tuned language model developed by Meta AI and released under an open-source license \cite{meta2025llama33}. The model is designed to deliver strong performance across complex reasoning, multilingual understanding, coding, and instruction-following tasks. Owing to its larger parameter scale, it consistently achieves competitive results on several reasoning and mathematical benchmarks while maintaining robust performance across multiple languages. In this study, the model was accessed through the Groq API to support efficient inference during large-scale evaluation. Llama-3.3-70B-Versatile was selected as a high-capacity dense transformer model with 70 billion parameters to serve as a performance benchmark. Including this model allows us to investigate how increased model capacity influences perturbated Bengali mathematical reasoning performance under different prompting strategies. It also serves as an important reference for comparing the performance of smaller models against a larger instruction-tuned LLM. As an open-source model freely available through the Groq API, it enables high-quality research without the financial constraints associated with proprietary models. Its proven performance on multilingual reasoning tasks makes it particularly suitable for evaluating Bengali mathematical reasoning, where linguistic complexity presents additional challenges for language models.

\subsubsection{Llama-4-Scout-17B-16E-Instruct}

Llama-4-Scout-17B-16E-Instruct is an open-source instruction-tuned language model developed by Meta AI based on a Mixture-of-Experts (MoE) architecture \cite{meta2025llama4}. Unlike conventional dense transformer models, the MoE architecture activates only a subset of experts during inference, improving computational efficiency while maintaining strong reasoning performance. The model is released under an open-source license and is designed to support multilingual understanding, long-context processing, and complex instruction-following tasks. In this study, Llama-4-Scout-17B-16E-Instruct was accessed through the Groq API and included as the representative MoE model among the evaluated language models. With 17 billion parameters and 16 experts, it represents an innovative architectural approach that balances computational efficiency with reasoning capability. The model was selected to examine whether sparse expert activation provides advantages over dense transformer architectures when solving perturbated Bengali mathematical reasoning problems. Comparing this model with dense LLMs enables a broader analysis of how different architectural designs influence reasoning accuracy, computational efficiency, and robustness. As an open-source model freely available through the Groq API, it allows researchers to explore the benefits of MoE architecture without the barriers of proprietary systems. Its inclusion in our study provides valuable insights into whether architectural innovations can compensate for smaller parameter counts in the context of Bengali perturbated mathematical reasoning.

\subsubsection{GPT-OSS-20B}

GPT-OSS-20B is an open-weight instruction-tuned large language model developed by OpenAI and released under an open-source license \cite{openai2025gptoss,bi2025gptoss}. The model is designed for advanced reasoning, instruction following, and general-purpose natural language understanding. Despite its moderate parameter size, the model demonstrates competitive performance across various reasoning tasks, including mathematical problem solving, logical inference, and multilingual language understanding. In this study, GPT-OSS-20B was accessed through the Groq API, which provides fast and reliable inference for large-scale evaluation. The model was selected as a medium-scale model with 20 billion parameters to investigate how model size influences perturbated Bengali mathematical reasoning performance under different prompting strategies. Its inclusion enables a direct comparison with the larger GPT-OSS-120B model while maintaining the same experimental environment, providing insights into the trade-off between computational efficiency and reasoning capability. As an open-source model freely available through the Groq API, it offers researchers a balance between performance and resource requirements. Evaluating GPT-OSS-20B alongside the selected Qwen and Llama models provides valuable insights into how different model families and training approaches influence mathematical reasoning in Bengali. The model's proven reasoning capabilities make it a strong candidate for evaluating the challenges of perturbated Bengali mathematical problems.

\subsubsection{GPT-OSS-120B}

GPT-OSS-120B is a high-capacity open-weight large language model developed by OpenAI and released under an open-source license \cite{openai2025gptoss120b,bi2025gptoss}. The model is designed for complex reasoning, multilingual understanding, and advanced instruction-following tasks. Owing to its substantially larger model capacity, GPT-OSS-120B is expected to provide stronger reasoning ability and better generalization on challenging mathematical problems than smaller language models. The model supports an extensive context window, enabling it to process long reasoning chains and complex problem formulations. In this study, the model was accessed through the Groq API, ensuring a consistent inference environment for all evaluated models. GPT-OSS-120B was selected as the largest model in our experiments with 120 billion parameters to examine the impact of model scale on perturbated Bengali mathematical reasoning performance. Its inclusion provides an upper-bound reference for comparing reasoning performance across different model families and prompting strategies. As an open-source model freely available through the Groq API, it represents the state-of-the-art in accessible large language models for academic research. By evaluating the largest available open-source model under the same experimental settings as the other selected LLMs, this study systematically analyzes how increased model capacity influences mathematical reasoning in perturbated Bengali. The model's high-capacity architecture makes it particularly suitable for handling the complex reasoning required by perturbated mathematical problems, where multiple reasoning steps and careful attention to linguistic nuances are essential for accurate problem solving.

\begin{table*}[htbp]
\centering
\caption{Model Specifications }
\label{tab:models}
\fontsize{9pt}{11pt}\selectfont
\renewcommand{\arraystretch}{1.70}
\setlength{\tabcolsep}{4.5pt}
\resizebox{\textwidth}{!}{%
\begin{tabular}{|>{\arraybackslash}p{4.5cm}|
                >{\centering\arraybackslash}p{2.2cm}|
                >{\centering\arraybackslash}p{2.8cm}|
                >{\centering\arraybackslash}p{1.2cm}|
                >{\centering\arraybackslash}p{2.2cm}|}
\hline

\centering \textbf{Model} &
\textbf{Parameters} &
\textbf{Context Length} &
\textbf{API} &
\textbf{Open Source} \\
\hline

Qwen3-32B & 32B & 128K & Groq & Yes \\
\hline
Llama-3.1-8B-Instant & 8B & 128K & Groq & Yes \\
\hline
Llama-3.3-70B-Versatile & 70B & 128K & Groq & Yes \\
\hline
Llama-4-Scout-17B-16E-Instruct & 17B & 256K & Groq & Yes \\
\hline
GPT-OSS-120B & 120B & 1M & Groq & Yes \\
\hline
GPT-OSS-20B & 20B & 128K & Groq & Yes \\
\hline
\end{tabular}%
}

\end{table*}

\subsection{Prompting Strategy}
\label{sec:prompting}

The design of effective prompting strategies is fundamental to evaluating the mathematical reasoning capabilities of Large Language Models (LLMs). Prior research has demonstrated that LLM performance is highly sensitive to prompt design, with variations in instruction formulation, formatting, and contextual cues significantly influencing model outputs \cite{wei2022chain, kojima2022zero}. To ensure a fair and consistent evaluation across all models in our study, we developed a standardized prompting framework that provides clear, structured, and unambiguous instructions for solving mathematical reasoning tasks. This methodological rigor is essential for attributing performance differences to the models' inherent reasoning abilities rather than to variations in prompt engineering. We employed identical prompts across all evaluated models, including Qwen3-32B, Llama-3.1-8B-Instant, Llama-3.3-70B-Versatile, Llama-4-Scout-17B-16E-Instruct, GPT-OSS-20B, and GPT-OSS-120B.

\subsubsection{Standard Prompting}

Standard prompting, also referred to as direct answering (DA), represents the most fundamental interaction paradigm with a Large Language Model (LLM). In this configuration, the prompt is formulated as a clear, unambiguous instruction or direct question, without explicitly guiding the model's internal reasoning process or providing any worked examples. The model is expected to generate a final response solely based on its pre-trained parametric knowledge and the immediate context provided within the prompt.

\begin{figure}[htbp]
    \centering
    \includegraphics[width=1\textwidth]{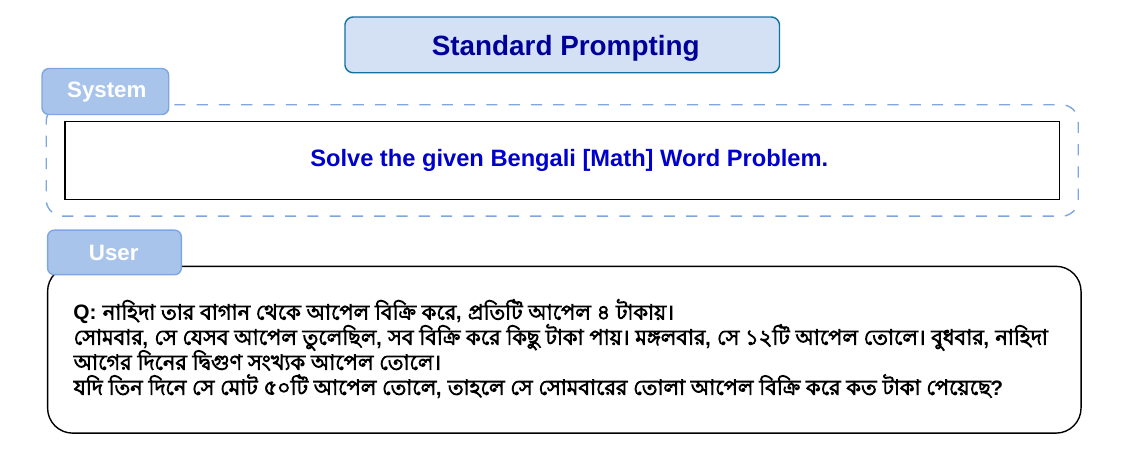}
    \caption{Illustration of the Standard prompting strategy}
    \label{fig:standard_example}
\end{figure}

This strategy serves a critical methodological function as the baseline condition in our comparative analysis. By establishing a performance benchmark under standard prompting, we create a control against which the efficacy of more sophisticated prompting techniques such as Chain-of-Thought can be quantified. Figure~\ref{fig:standard_example} provides an illustration of the Standard prompting strategy. Previous studies, including the work of Paul et al., have employed this structure to demonstrate the performance ceiling of LLMs when no explicit reasoning scaffolding is provided \cite{paul2025leveraging}. This approach is efficient in terms of token usage and computational cost, making it suitable for straightforward tasks where the model's inherent knowledge is sufficient. Its primary limitation, however, is that it does not scaffold the model's internal computation for complex, multi-step problems, making it vulnerable to errors on tasks requiring intricate reasoning.

\subsubsection{Chain-of-Thought (CoT) Prompting}

\begin{figure}[htbp]
    \centering
    \includegraphics[width=1\textwidth]{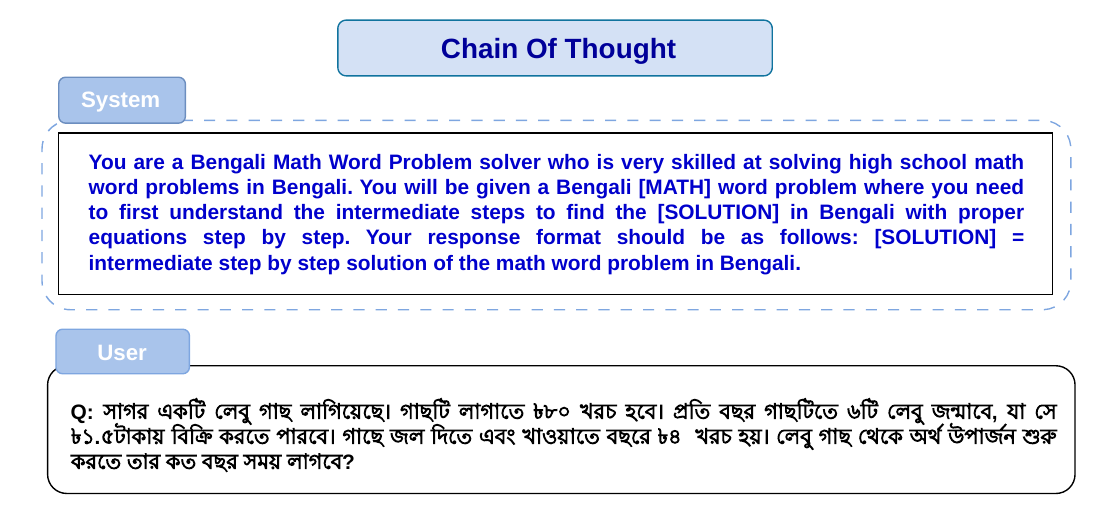}
    \caption{Illustration of the Chain-of-Thought prompting strategy, demonstrating the step-by-step reasoning process generated by the LLM.}
    \label{fig:cot_example}
\end{figure}

Chain-of-Thought (CoT) prompting is an advanced technique designed to enhance the model's performance on complex reasoning tasks by eliciting a structured sequence of intermediate reasoning steps prior to the final answer. This method, initially formalized by Wei et al., compels the model to decompose a problem into smaller, manageable sub-problems and address them sequentially, thereby mimicking human cognitive processes \cite{wei2022chain}. The prompt explicitly instructs the model to "think step by step," requiring the generation of a transparent reasoning pathway that logically connects the initial query to the final conclusion. In our implementation, we adopt the zero-shot variant, wherein the trigger phrase "Let's think step by step" is appended to the task instruction without the provision of any example rationales. Figure~\ref{fig:cot_example} provides an illustration of the Chain-of-Thought prompting strategy, demonstrating the step-by-step reasoning process generated by the LLM.

The primary justification for incorporating CoT prompting is its proven efficacy in improving performance on tasks that necessitate mathematical, logical, and symbolic reasoning \cite{kojima2022zero}. By externalizing the reasoning chain, the model reduces the likelihood of logical errors and enhances the interpretability and transparency of its output. Our methodology leverages this strategy to investigate the model's ability to perform execution steps the process of performing computation and symbolic manipulation which represents a critical bottleneck in standard prompting scenarios. This approach is particularly valuable in our experimental design, as it allows us to assess whether the LLM's performance on complex reasoning tasks can be improved through structured prompting, a finding consistently reported in prior work \cite{paul2025leveraging, wei2022chain}. It is important to note, however, that the utility of CoT prompting may be domain-specific; studies indicate that its benefits are most pronounced in symbolic reasoning tasks and may not consistently generalize to non-symbolic or commonsense reasoning problems.

\section{Results Analysis}

This section presents the performance evaluation of six Large Language Models (LLMs) across eight distinct perturbation types designed to test reasoning robustness. We analyze the models' accuracy under two prompting strategies: Chain-of-Thought (CoT) and Standard Prompting.  A total of 9000 samples were taken for the evaluation, where 1000 samples were from seed questions and the other 8000 are the variations of them. The analysis aims to identify which models exhibit superior reasoning capabilities and which perturbation types pose the greatest challenge.

\subsection{Performance Overview: Seed Questions}

To establish a baseline for comparison, we first evaluate model performance on unperturbed seed questions. Table~\ref{tab:seed_results} presents the accuracy percentages for all models under both prompting strategies. Parenthetical values indicate the performance change with CoT relative to Standard prompting, with $\uparrow$ denoting improvement and $\downarrow$ denoting decline. The highest accuracy in each row is bolded.

\begin{table}[!htb]
    \centering
    \setlength{\tabcolsep}{4.5pt} 
    \renewcommand{\arraystretch}{1.3}

    \renewcommand{\arraystretch}{1.5}
    \caption{Model Accuracy on Seed Questions (\%) Values in parentheses for CoT indicate performance change relative to Standard prompting, with $\uparrow$ denoting improvement and $\downarrow$ denoting decline.}
    \label{tab:seed_results}
    \begin{tabular}{lcccccc}
        \toprule
        \textbf{Prompting Strategy} & \textbf{Qwen3-32B} & \textbf{Llama-3.1} & \textbf{Llama-3.3} & \textbf{Llama-4-Scout} & \textbf{GPT-OSS} & \textbf{GPT-OSS} \\
        & & \textbf{8B-Instant} & \textbf{70B-Versatile} & \textbf{17B-16E-Instruct} & \textbf{120B} & \textbf{20B} \\
        \midrule
        Standard Prompting  & 13.98 & 90.15 & 95.90 & 95.68 & 88.03 & \textbf{96.08} \\
        Chain-of-Thought    & 76.25\textsubscript{($\uparrow$62.27)} & 66.88\textsubscript{($\downarrow$23.27)} & 90.25\textsubscript{($\downarrow$5.65)} & 90.55\textsubscript{($\downarrow$5.13)} & \textbf{92.70}\textsubscript{($\uparrow$4.67)} & 87.80\textsubscript{($\downarrow$8.28)} \\
        \bottomrule
    \end{tabular}
\end{table}

The seed question results reveal several important patterns. Under Standard prompting, most models demonstrate strong performance, with GPT-OSS-20B achieving the highest accuracy (96.08\%), closely followed by Llama-3.3-70B-Versatile (95.90\%) and Llama-4-Scout (95.68\%). However, Qwen3-32B exhibits exceptionally poor performance (13.98\%), suggesting a significant deficiency in its inherent reasoning capabilities for the given tasks.

Under CoT prompting, performance improves substantially for most models. GPT-OSS-120B achieves the highest accuracy (92.70\%), followed by Llama-3.3-70B-Versatile (90.25\%) and Llama-4-Scout (90.55\%). Notably, Qwen3-32B shows the largest improvement (from 13.98\% to 76.25\%), indicating that CoT prompting is particularly beneficial for models with weaker inherent reasoning capabilities. Conversely, Llama-3.1-8B-Instant shows a decline in performance under CoT (from 90.15\% to 66.88\%), suggesting that this model may not be well-aligned with the CoT prompting format for seed questions.

\subsection{Performance Overview: Standard Prompting}

Table~\ref{tab:standard_results} presents the accuracy percentages for each model on perturbated questions under Standard prompting. The highest accuracy in each row is bolded.

\begin{table}[!htb]
    \centering
    \setlength{\tabcolsep}{4.5pt} 
    \renewcommand{\arraystretch}{1.3}

    \renewcommand{\arraystretch}{1.5}
    \caption{Model Accuracy on Perturbated Questions with Standard Prompting (\%).}
    \label{tab:standard_results}
    \begin{tabular}{lcccccc}
        \toprule
        \textbf{Perturbation Type} & \textbf{Qwen3-32B} & \textbf{Llama-3.1} & \textbf{Llama-3.3} & \textbf{Llama-4-Scout} & \textbf{GPT-OSS} & \textbf{GPT-OSS} \\
        & & \textbf{8B-Instant} & \textbf{70B-Versatile} & \textbf{17B-16E-Instruct} & \textbf{120B} & \textbf{20B} \\
        \midrule
        Numerical Substitution  & 42.91 & 67.66 & 81.84 & \textbf{85.23} & 84.03 & 79.64 \\
        Digit Expansion         & 39.20 & 61.40 & 81.40 & 81.60 & \textbf{82.40} & 82.60 \\
        Integer-Decimal- & 24.50 & 57.63 & 76.71 & \textbf{78.11} & 78.11 & 74.50 \\
        \hspace{0 cm}Fraction Conversion & & & & & & \\
        Adding Operation        & 21.24 & 53.31 & 68.94 & \textbf{74.55} & 72.14 & 68.34 \\
        Reversing Operation     & 35.60 & 67.60 & 82.40 & \textbf{86.00} & 86.00 & 82.40 \\
        Problem Understanding   & 47.50 & 71.86 & 85.43 & 87.43 & \textbf{89.82} & 87.62 \\
        Distraction Insertion   & 33.53 & 58.88 & 80.64 & 79.44 & \textbf{83.83} & 75.85 \\
        Critical Thinking       & 6.60  & 14.20 & \textbf{31.80} & 21.40 & 28.60 & 25.20 \\
        \bottomrule
    \end{tabular}
\end{table}

Under Standard prompting, the Llama-4-Scout and GPT-OSS-120B models demonstrate the strongest performance across most perturbation types, consistently achieving accuracy rates exceeding 80\% on simpler perturbations. The Llama-3.3-70B-Versatile model also performs competitively, while Qwen3-32B exhibits significantly lower performance, with accuracy rates dropping below 25\% on more challenging perturbations.

The \textit{Critical Thinking} perturbation proves to be the most challenging across all models, with accuracy rates ranging from a low of 6.60\% (Qwen3-32B) to a high of only 31.80\% (Llama-3.3-70B). This suggests that current LLMs lack the fundamental reasoning capabilities required for complex, multi-step logical problems. The \textit{Adding Operation} and \textit{Integer-Decimal-Fraction Conversion} perturbations also present significant challenges, with accuracy rates consistently below 80\% for most models.

The \textit{Problem Understanding} perturbation yields the highest accuracy across models, with GPT-OSS-120B achieving 89.82\% and Llama-4-Scout achieving 87.43\%. The \textit{Reversing Operation} and \textit{Numerical Substitution} perturbations also show strong performance, with top models exceeding 85\% accuracy. This indicates that models possess relatively robust syntactic and semantic processing capabilities.

\subsection{Performance Overview: CoT Analysis}

Table~\ref{tab:cot_results} presents the accuracy percentages for each model on perturbated questions under Chain-of-Thought prompting. Parenthetical values indicate the performance change relative to Standard prompting, with $\uparrow$ denoting improvement and $\downarrow$ denoting decline.

\begin{table}[!htb]
    \centering
    \setlength{\tabcolsep}{4.4pt} 
    \renewcommand{\arraystretch}{1.3}

    \renewcommand{\arraystretch}{1.5}
    \caption{Model Accuracy on Perturbated Questions with CoT Prompting (\%) Values in parentheses indicate performance change relative to Standard prompting, with $\uparrow$ denoting improvement and $\downarrow$ denoting decline ≈ denotes negligible change.}
    \label{tab:cot_results}
    \begin{tabular}{lcccccc}
        \toprule
        \textbf{Perturbation Type} & \textbf{Qwen3-32B} & \textbf{Llama-3.1} & \textbf{Llama-3.3} & \textbf{Llama-4-Scout} & \textbf{GPT-OSS} & \textbf{GPT-OSS} \\
        & & \textbf{8B-Instant} & \textbf{70B-Versatile} & \textbf{17B-16E-Instruct} & \textbf{120B} & \textbf{20B} \\
        \midrule
        Numerical Substitution  & 68.06\textsubscript{($\uparrow$25.15)} & 66.27\textsubscript{($\downarrow$1.39)} & 87.03\textsubscript{($\uparrow$5.19)} & \textbf{88.42}\textsubscript{($\uparrow$3.19)} & 83.23\textsubscript{($\downarrow$0.80)} & 81.64\textsubscript{($\uparrow$2.00)} \\
        Digit Expansion         & 67.80\textsubscript{($\uparrow$28.60)} & 58.00\textsubscript{($\downarrow$3.40)} & 85.20\textsubscript{($\uparrow$3.80)} & 84.20\textsubscript{($\uparrow$2.60)} & 80.40\textsubscript{($\downarrow$2.00)} & \textbf{81.40}\textsubscript{($\downarrow$1.20)} \\
        Integer-Decimal- & 55.02\textsubscript{($\uparrow$30.52)} & 52.61\textsubscript{($\downarrow$5.02)} & 79.12\textsubscript{($\uparrow$2.41)} & \textbf{79.52}\textsubscript{($\uparrow$1.41)} & 75.90\textsubscript{($\downarrow$2.21)} & 73.29\textsubscript{($\downarrow$1.21)} \\
        \hspace{0cm}Fraction Conversion & & & & & & \\
        Adding Operation        & 45.49\textsubscript{($\uparrow$24.25)} & 50.10\textsubscript{($\downarrow$3.21)} & \textbf{76.75}\textsubscript{($\uparrow$7.81)} & 75.95\textsubscript{($\uparrow$1.40)} & 70.54\textsubscript{($\downarrow$1.60)} & 65.53\textsubscript{($\downarrow$2.81)} \\
        Reversing Operation     & 63.40\textsubscript{($\uparrow$27.80)} & 68.00\textsubscript{($\uparrow$0.40)} & 87.00\textsubscript{($\uparrow$4.60)} & \textbf{88.00}\textsubscript{($\uparrow$2.00)} & 86.60\textsubscript{($\uparrow$0.60)} & 81.60\textsubscript{($\downarrow$0.80)} \\
        Problem Understanding   & 77.05\textsubscript{($\uparrow$29.55)} & 69.06\textsubscript{($\downarrow$2.80)} & 89.22\textsubscript{($\uparrow$3.79)} & 88.82\textsubscript{($\uparrow$1.39)} & \textbf{91.82}\textsubscript{($\uparrow$2.00)} & 86.63\textsubscript{($\downarrow$0.99)} \\
        Distraction Insertion   & 54.09\textsubscript{($\uparrow$20.56)} & 59.28\textsubscript{($\uparrow$0.40)} & \textbf{84.43}\textsubscript{($\uparrow$3.79)} & 81.44\textsubscript{($\uparrow$2.00)} & 84.43\textsubscript{($\uparrow$0.60)} & 77.84\textsubscript{($\uparrow$1.99)} \\
        Critical Thinking       & 11.00\textsubscript{($\uparrow$4.40)} & 16.40\textsubscript{($\uparrow$2.20)} & 31.80\textsubscript{($\approx$0.00)} & 27.40\textsubscript{($\uparrow$6.00)} & 31.80\textsubscript{($\uparrow$3.20)} & \textbf{33.00}\textsubscript{($\uparrow$7.80)} \\
        \bottomrule
    \end{tabular}
\end{table}

CoT prompting substantially improves model performance across all perturbation types. The Llama-3.3-70B-Versatile and Llama-4-Scout models achieve the highest overall accuracy, consistently exceeding 85\% on most perturbations. The GPT-OSS-120B model also performs competitively, while Qwen3-32B and Llama-3.1-8B-Instant show moderate improvements but remain the weakest performers. Under CoT, Qwen3-32B achieves the lowest average accuracy (54.28\%), closely followed by Llama-3.1-8B-Instant (53.38\%).

The most substantial improvements from CoT prompting are observed in arithmetic and numeric perturbations, particularly for Qwen3-32B. Qwen3-32B shows dramatic gains across all perturbation types, with improvements ranging from +20.56 (Distraction Insertion) to +30.52 (Integer-Decimal-Fraction Conversion) percentage points. This confirms that explicit step-by-step reasoning instructions effectively guide models through arithmetic and logical steps disrupted by numeric perturbations. The average improvement of +23.85 percentage points for Qwen3-32B suggests that CoT prompting is particularly beneficial for models with weaker inherent reasoning capabilities.

In contrast, the larger models show more modest improvements. Llama-3.3-70B-Versatile and Llama-4-Scout achieve average gains of +3.92 and +2.50 percentage points, respectively, indicating that these models already possess strong inherent reasoning capabilities. Interestingly, GPT-OSS-120B shows a marginal decline (-0.03 percentage points average), suggesting the model may be optimized for direct answer generation rather than step-by-step reasoning. Llama-3.1-8B-Instant demonstrates the most concerning pattern, with an average decline of -1.60 percentage points, suggesting the model is not well-aligned with the CoT prompting format.

Despite CoT prompting, the \textit{Critical Thinking} perturbation remains exceptionally challenging, with accuracy rates ranging from 11.00\% (Qwen3-32B) to 33.00\% (GPT-OSS-20B). The minimal improvement from CoT (0.00 to 7.80 percentage points) suggests that current LLMs fundamentally lack the capabilities required for complex logical reasoning tasks. The Adding Operation perturbation also continues to pose difficulties, with accuracy rates below 77\% for all models. This indicates that while CoT prompting is beneficial for many tasks, it cannot fully compensate for fundamental reasoning deficiencies in current LLM architectures.

(Numerical Substitution, Digit Expansion, Integer-Decimal-Fraction Conversion, Adding Operation) consistently yield the lowest accuracy rates across both prompting strategies. The Llama-4-Scout model demonstrates the strongest performance on these perturbations, achieving the highest scores on \textit{Numerical Substitution} (88.42\%), \textit{Digit Expansion} (84.20\%), and \textit{Integer-Decimal-Fraction Conversion} (79.52\%) under CoT. However, all models struggle with \textit{Adding Operation}, suggesting fundamental limitations in arithmetic reasoning.

(Reversing Operation, Problem Understanding) demonstrate the highest resilience across all models, with CoT accuracy rates exceeding 86\% for top performers. \textit{Problem Understanding} achieves the highest accuracy, with GPT-OSS-120B reaching 91.82\% under CoT. The high baseline Standard performance (71.86\% to 89.82\%) suggests models possess relatively robust syntactic and semantic processing capabilities.

(Distraction Insertion) shows moderate improvement with CoT, with gains ranging from 0.40 to 20.56 percentage points. GPT-OSS-120B and Llama-3.3-70B achieve the highest performance under CoT (84.43\% each). However, Standard performance (33.53\% to 83.83\%) indicates models often get misled by inserted distractions, and CoT helps systematically evaluate and disregard irrelevant content.

(Critical Thinking) shows uniformly poor performance across all models and prompting strategies. Even the best-performing models achieve only 31.80\% to 33.00\% under CoT, with minimal improvement from Standard (6.60\% to 31.80\%). This indicates that while CoT prompting is effective for many tasks, it cannot compensate for missing reasoning abilities required for complex logical problems.

\subsection{Detailed Model-wise Analysis}

\subsubsection{Llama-3.3-70B-Versatile: The Overall Best Performer}

The Llama-3.3-70B-Versatile model demonstrates the highest overall performance under CoT prompting, achieving accuracy rates exceeding 85\% on most perturbation types. The model achieves its highest performance on \textit{Problem Understanding} (89.22\%), \textit{Reversing Operation} (87.00\%), and \textit{Numerical Substitution} (87.03\%). Its performance on seed questions (90.25\% under CoT) establishes a strong baseline.

However, the model shows notable weaknesses in \textit{Adding Operation} (76.75\%) and \textit{Critical Thinking} (31.80\%). The significant drop in performance on arithmetic tasks indicates that even state-of-the-art models struggle with mathematical reasoning when numbers are presented in perturbated formats. The CoT prompting provides limited improvement on \textit{Critical Thinking} (0.00 percentage points), confirming that complex logical reasoning remains a significant challenge.

\subsubsection{Llama-4-Scout-17B-16E-Instruct: Strong and Consistent Performer}

The Llama-4-Scout model achieves performance comparable to Llama-3.3-70B, with accuracy rates exceeding 84\% on most perturbations. Its highest accuracy is achieved on \textit{Numerical Substitution} (88.42\%), \textit{Problem Understanding} (88.82\%), and \textit{Reversing Operation} (88.00\%). The model demonstrates particular strength in handling numeric format conversions, achieving 79.52\% on \textit{Integer-Decimal-Fraction Conversion}.

Similar to other models, Llama-4-Scout struggles with \textit{Adding Operation} (75.95\%) and \textit{Critical Thinking} (27.40\%). The model shows moderate improvement with CoT (average +2.50 percentage points), suggesting that it is well-aligned with explicit reasoning instructions but already possesses strong inherent capabilities.

\subsubsection{GPT-OSS-120B: Strong Inherent Capabilities}

The GPT-OSS-120B model achieves 92.70\% accuracy on seed questions under CoT, the highest among all models. On perturbated questions, the model excels at \textit{Problem Understanding} (91.82\%) and \textit{Reversing Operation} (86.60\%). However, its performance on arithmetic perturbations is notably lower, with \textit{Adding Operation} at 70.54\% and \textit{Numerical Substitution} at 83.23\%.

Interestingly, GPT-OSS-120B shows a slight decline in performance with CoT prompting on several perturbation types, including \textit{Numerical Substitution} (-0.80\%), \textit{Digit Expansion} (-2.00\%), and \textit{Integer-Decimal-Fraction Conversion} (-2.21\%). This suggests that the model may have been optimized for direct answer generation and does not benefit from explicit reasoning instructions.

\subsubsection{GPT-OSS-20B: Competitive Mid-tier Performer}

The GPT-OSS-20B model achieves competitive performance on seed questions (87.80\% under CoT). On perturbated questions, the model achieves its highest accuracy on \textit{Problem Understanding} (86.63\%) and \textit{Numerical Substitution} (81.64\%). The model shows modest improvement with CoT (average +0.60 percentage points) and achieves the highest \textit{Critical Thinking} performance under CoT (33.00\%), surpassing all other models.

\subsubsection{Qwen3-32B: Most Improved with CoT}

Qwen3-32B exhibits the most dramatic improvement with CoT prompting, with an average gain of 23.85 percentage points. On seed questions, the model improves from 13.98\% (Standard) to 76.25\% (CoT), representing a 62.27 percentage point increase. This suggests that Qwen3-32B possesses latent reasoning capabilities that can be effectively unlocked through explicit prompting.

However, even with CoT, Qwen3-32B remains the weakest performer on most perturbations, with accuracy rates below 70\% on all tasks except \textit{Problem Understanding} (77.05\%) and \textit{Numerical Substitution} (68.06\%). The model struggles particularly with \textit{Critical Thinking} (11.00\%), \textit{Adding Operation} (45.49\%), and \textit{Distraction Insertion} (54.09\%).

\subsubsection{Llama-3.1-8B-Instant: Mixed Performance}

Llama-3.1-8B-Instant shows a decline in performance with CoT prompting on most perturbation types, with an average decrease of 1.60 percentage points. On seed questions, the model drops from 90.15\% (Standard) to 66.88\% (CoT), representing a 23.27 percentage point decline. This suggests that the model is not well-aligned with the CoT prompting format.

However, the model achieves competitive performance on some perturbations under CoT, including \textit{Problem Understanding} (69.06\%) and \textit{Reversing Operation} (68.00\%). The model's performance on \textit{Critical Thinking} is the lowest among all models under CoT (16.40\%), indicating significant limitations in complex reasoning.

\subsection{Seed vs. Perturbated Performance Analysis}

To understand the impact of perturbations on model performance, we compare accuracy on seed questions against perturbated questions. Table~\ref{tab:seed_perturbation_gap} presents the performance gap between seed and perturbated questions.

\begin{table}[!htb]
    \centering
    \setlength{\tabcolsep}{5pt} 
    \renewcommand{\arraystretch}{1.3}
    \caption{Performance Gap: Seed vs. Perturbated Questions (\%) The gap represents the difference between seed question accuracy and average perturbated question accuracy. Negative values indicate performance degradation on perturbated questions.}
    \label{tab:seed_perturbation_gap}
    \begin{tabular}{lcccccc}
        \toprule
        \textbf{Prompting Strategy} & 
        \makecell{\textbf{Qwen3-}\\ \textbf{32B}} & 
        \makecell{\textbf{Llama-3.1}\\ \textbf{8B-Instant}} & 
        \makecell{\textbf{Llama-3.3}\\ \textbf{70B-Versatile}} & 
        \makecell{\textbf{Llama-4-Scout}\\ \textbf{17B-16E-Instruct}} & 
        \makecell{\textbf{GPT-OSS}\\ \textbf{120B}} & 
        \makecell{\textbf{GPT-OSS}\\ \textbf{20B}} \\
        \midrule
        Standard Prompting & -28.61 & -39.92 & -25.26 & -22.85 & \textbf{-20.22} & -27.66 \\
        Chain-of-Thought   & -20.26 & -12.70 & -12.68 & -13.81 & -14.74 & \textbf{-11.36} \\
        \bottomrule
    \end{tabular}
\end{table}

The seed-perturbation gap reveals several important patterns. Under Standard prompting, all models show substantial performance degradation on perturbated questions, with average gaps ranging from 20.22 to 39.92 percentage points. The Llama-3.1-8B-Instant shows the largest gap (39.92\%), followed by Qwen3-32B (28.61\%).

Under CoT prompting, the performance gap narrows considerably for most models. The largest improvement is observed for Qwen3-32B, which reduces its gap from 28.61 to 20.26 percentage points. The Llama-3.1-8B-Instant shows the most dramatic reduction, from 39.92 to 12.70 percentage points. This suggests that CoT prompting helps models better handle perturbations by providing structured reasoning frameworks.

The GPT-OSS models show the smallest gaps under both prompting strategies, indicating greater robustness to perturbations. The GPT-OSS-120B achieves the smallest gap under Standard (20.22\%), while GPT-OSS-20B achieves the smallest gap under CoT (11.36\%).

\subsection{Discussion}

The experimental results reveal several critical insights into the reasoning robustness of contemporary Large Language Models. The differential impact of Chain-of-Thought prompting across models is particularly noteworthy weaker models like Qwen3-32B show dramatic improvements (averaging +23.85 percentage points), suggesting they possess latent reasoning abilities that remain dormant without explicit guidance. In contrast, larger models like Llama-3.3-70B and Llama-4-Scout demonstrate strong inherent robustness with only modest CoT gains, while GPT-OSS-120B and Llama-3.1-8B-Instant surprisingly show performance declines, indicating that not all models benefit uniformly from reasoning instructions and highlighting the importance of model-specific prompting alignment. Persistent challenges in arithmetic reasoning across all models represent a fundamental bottleneck, with perturbations like Adding Operation and Integer-Decimal-Fraction Conversion consistently yielding the lowest accuracy rates below 80\% even under CoT. This systematic failure suggests LLMs rely on surface-level pattern recognition rather than genuine mathematical understanding, with tokenization and attention mechanisms prioritizing linguistic patterns over numerical precision a concerning limitation for applications requiring reliable computation.

Most alarmingly, the uniformly poor performance on Critical Thinking perturbations (accuracy below 34\% across all models) with minimal CoT improvement (0.00-7.80 percentage points) reveals that current LLMs fundamentally lack complex, multi-step logical reasoning capabilities. This limitation persists regardless of prompting strategy, indicating architectural constraints rather than prompt engineering issues, with profound implications for deployment in legal, medical, or strategic domains requiring genuine reasoning beyond pattern matching.

The seed versus perturbated performance gap analysis shows substantial degradation under standard prompting (20.22-39.92 percentage points), confirming high sensitivity to surface-level input variations. While CoT narrows this gap, persistent degradation suggests models rely on format-specific patterns rather than true reasoning principles. GPT-OSS models demonstrate the greatest robustness, maintaining the smallest performance gaps and suggesting that training diversity improves perturbation resistance. Syntactic and structural transformations show the highest resilience, with Problem Understanding reaching 91.82\% under CoT, indicating robust semantic processing. However, arithmetic and higher-order reasoning tasks expose fundamental fragility, with implications for model selection Llama-3.3-70B and Llama-4-Scout are recommended for robust reasoning applications, while Qwen3-32B may serve as a cost-effective option when CoT can be applied. The persistent challenges in arithmetic and critical thinking underscore urgent research directions, including specialized arithmetic training, novel reasoning frameworks, and architectural innovations that support genuine logical reasoning rather than pattern recognition.

\section{Limitations}

While this study provides the first comprehensive perturbed mathematical reasoning benchmark for Bangla, several limitations should be acknowledged.

First, our evaluation is restricted to six open-source models due to budgetary constraints. Proprietary models such as GPT-4, Claude, and Gemini were not included, which may limit the generalizability of our findings to state-of-the-art commercial systems. Future work should expand the model roster to include both proprietary and domain-specialized models. Second, we employ only two prompting standard prompting and chain-of-thought prompting. More advanced techniques such as few-shot prompting, self-consistency, Tree-of-Thought, and Program-of-Thought were not explored. A broader investigation of prompting variants could reveal optimal strategies for Bangla mathematical reasoning. Third, the perturbation types evaluated while systematically designed represent only a subset of possible input variations. Real-world user queries exhibit a wider range of linguistic diversity including misspellings, code-switching, and colloquial expressions not captured here.

Finally, our primary evaluation metric is exact match accuracy, which does not capture partial correctness or reasoning quality. Finer-grained error analysis would provide deeper insights into model failure modes.

\section{Future Work}

Despite these limitations, GSM-PLUS-BN serves as a valuable foundation for future research. First, we plan to expand our evaluation to include proprietary models (GPT-4, Claude, Gemini) and domain-specialized reasoning models, as our current study was limited to open-source models due to budgetary constraints. Second, we intend to explore a broader range of prompting strategies, including few-shot prompting, self-consistency, Tree-of-Thought, Program-of-Thought, and cross-lingual prompting, to identify optimal approaches for Bangla reasoning tasks. Finally, longitudinal studies tracking model performance over time will reveal whether robustness gaps for low-resource languages are narrowing or widening.

\section{Conclusion}

In this study, we evaluated six Large Language Models across eight perturbation types Numerical Substitution, Digit Expansion, Integer-Decimal-Fraction Conversion, Adding Operation, Reversing Operation, Problem Understanding, Distraction Insertion, and Critical Thinking designed to systematically test reasoning robustness from our proposed dataset (GSM-PLUS-BN). We employed Standard prompting and Chain-of-Thought prompting strategies to evaluate model performance, with GPT-OSS-120B achieving the highest seed question accuracy at 92.70\%, while Llama-3.3-70B-Versatile emerged as the most consistent performer on perturbated questions, exceeding 85\% accuracy on most perturbations under CoT. Qwen3-32B showed the most dramatic improvement with CoT (average +23.85 percentage points), whereas Llama-3.1-8B-Instant demonstrated performance degradation under CoT, highlighting model-specific prompting sensitivities. Critical Thinking perturbations proved universally challenging, with accuracy remaining below 34\% across all models, revealing fundamental limitations in complex logical reasoning. Arithmetic perturbations consistently yielded the lowest accuracy, indicating that current LLMs rely on pattern recognition rather than genuine mathematical understanding. CoT prompting effectively narrowed the seed-perturbation performance gap but could not fully compensate for architectural limitations in mathematical and logical reasoning. These findings provide practical guidance for model selection while highlighting critical directions for advancing LLM reasoning capabilities.

\bibliographystyle{unsrt}  
\bibliography{references}  


\end{document}